\documentclass[conference]{IEEEtran}
\IEEEoverridecommandlockouts

\usepackage{cite}
\usepackage{amsmath,amssymb,amsfonts,bm}
\usepackage{graphicx}
\usepackage{textcomp}
\usepackage{xcolor}
\usepackage{booktabs}
\usepackage{multirow}
\usepackage{url}

\begin{document}

\title{Cross-lingual Relation Extraction with Large Language Models: Zero-Shot, Few-Shot, and Fine-Tuned Evaluation on Romanian}

\author{
\IEEEauthorblockN{Dragoș-Mitruț Vasile\IEEEauthorrefmark{1}, Elena-Simona Apostol\IEEEauthorrefmark{1}, Ștefan-Adrian Toma\IEEEauthorrefmark{2}, Adrian Paschke\IEEEauthorrefmark{3}\IEEEauthorrefmark{4}, Ciprian-Octavian Truic\u{a}\IEEEauthorrefmark{1}\IEEEauthorrefmark{5}}
\IEEEauthorblockA{\IEEEauthorrefmark{1}National University of Science and Technology POLITEHNICA Bucharest,\\
Splaiul Independenței 313, București 060042, Romania}
\IEEEauthorblockA{\IEEEauthorrefmark{2}
Military Technical Academy `Ferdinand I',
Bulevardul George Coșbuc 39-49, 050141, Bucharest, Romania
}
\IEEEauthorblockA{\IEEEauthorrefmark{3}Freie Universität Berlin, Arnimallee 14, Berlin, 14195, Germany}
\IEEEauthorblockA{\IEEEauthorrefmark{4}Fraunhofer Institute for Open Communication Systems, Kaiserin-Augusta-Allee 31, Berlin, 10589, Germany}
\IEEEauthorblockA{\IEEEauthorrefmark{5}Academy of Romanian Scientists, Ilfov 3, Bucharest, 050044, Romania\\
Email: dragos.vasile2603@upb.ro, elena.apostol@upb.ro, stefan.toma@mta.ro, paschke@inf.fu-berlin.de, ciprian.truica@upb.ro}
}

\maketitle

\begin{abstract}
Relation extraction (RE) for low-resource languages is typically constrained by the lack of annotated corpora.
We investigate the feasibility of cross-lingual RE for Romanian by combining automatic dataset translation with large language model (LLM) inference.
We translate the SemEval-2010 Task~8 benchmark from English to Romanian using an LLM-based translation pipeline and evaluate Gemma~4 31B under zero-shot, few-shot, and QLoRA fine-tuned configurations, against four encoder baselines spanning 125M to 560M parameters: XLM-RoBERTa (base and large), Romanian BERT, and RoBERT-large.
We assess two task formulations: relation classification with marked entities and end-to-end extraction.
Our results show that Romanian incurs a 3 to 5 percentage point (pp) drop relative to English in prompt-only settings, that few-shot prompting provides marginal gains over zero-shot, and that QLoRA fine-tuning improves macro F1-Score by more than 22 percentage points in both languages while reducing the cross-lingual gap from 3.3 to 1.4pp.
The encoder baselines come within 1--4pp of QLoRA Gemma on Romanian despite being 50--250 times smaller, with monolingual Romanian BERT at 125M parameters matching multilingual XLM-R at 278M.
The case for using a 31B model for single-task RE on Romanian is therefore weak in deployment scenarios where compute matters.
We release the translated dataset, evaluation code, and trained models.
\end{abstract}

\begin{IEEEkeywords}
relation extraction, cross-lingual NLP, large language models, Romanian, few-shot learning, QLoRA
\end{IEEEkeywords}

\section{Introduction}

Relation extraction (RE) is the task of identifying semantic relations between entities mentioned in text.
While substantial progress has been made for English, low-resource languages such as Romanian remain largely unexplored due to the scarcity of annotated datasets.
Building such resources from scratch requires trained annotators and careful guideline design, both of which are expensive and time-consuming.

An alternative approach is to transfer existing English benchmarks to the target language through automatic translation.
This raises several research questions:
\begin{itemize}
    \item[$\bm{Q_1}$] How much performance is lost through translation artifacts? 
    \item[$\bm{Q_2}$] Can LLM-based zero-shot and few-shot inference bridge the gap?
    \item[$\bm{Q_3}$] To what extent can parameter-efficient fine-tuning on translated data improve results?
\end{itemize}

We address these questions using SemEval-2010 Task~8~\cite{hendrickx2010semeval}, a well-established RE benchmark with 10 relation types and directional labels.
We translate the entire dataset from English to Romanian using Claude Haiku~\cite{anthropic2024claude} and evaluate Gemma~4 31B-it~\cite{gemma2026}, a recent open-weight instruction-tuned model, under three prompting configurations.
We further fine-tune the model using QLoRA~\cite{dettmers2023qlora} on the translated training set to measure the effect of domain adaptation, and compare against four encoder baselines from 125M to 560M parameters trained on the same data.

Our contributions are as follows:
\begin{itemize}
    \item[$\bm{C_1}$] We construct and validate a Romanian version of SemEval-2010 Task~8 through LLM-based translation with automatic quality checks.
    \item[$\bm{C_2}$] We provide a systematic comparison of zero-shot, few-shot (1, 3, 5 examples), and fine-tuned LLM performance on both the original English and translated Romanian data, with four encoder baselines from 125M to 560M parameters for context.
    \item[$\bm{C_3}$] We evaluate two task formulations (classification with given entities and end-to-end extraction) and analyze the specific challenges of each for cross-lingual transfer.
\end{itemize}

We make the translated dataset\footnote{\url{https://huggingface.co/datasets/DS4AI-UPB/romanian-re-semeval}}, evaluation code\footnote{\url{https://github.com/DS4AI-UPB/crosslingual-romanian-re}}, and trained models\footnote{\url{https://huggingface.co/DS4AI-UPB}} publicly available.

The remainder of the paper is organized as follows.
Section~\ref{sec:related} reviews prior work on relation extraction, cross-lingual transfer, and LLM-based information extraction.
Section~\ref{sec:method} describes the dataset construction process, the two task formulations, and the three inference configurations.
Section~\ref{sec:results} reports the experimental results and discusses the cross-lingual gap, the effect of few-shot examples, and the difficulty of end-to-end extraction.
Section~\ref{sec:future} discusses limitations and future research directions.

\section{Related Work}
\label{sec:related}

\noindent \textbf{Relation Extraction.} 
Early work on RE used handcrafted patterns and tree kernels.
Zelenko et al.~\cite{zelenko2003kernel} apply kernel methods to news articles. 
Convolutional networks with position embeddings, introduced by Zeng et al.~\cite{zeng2014relation}, became the standard neural baseline on SemEval-2010 Task 8.
Pretrained transformer encoders followed, with both span-based classifiers~\cite{wadden2019entity} and entity-marker pretraining~\cite{baldini2019relation} reporting strong numbers on this benchmark.
Sequence-to-sequence formulations of RE~\cite{paolini2021structured} treat the output as augmented natural language and are conceptually close to our end-to-end extraction setting.\\

\noindent \textbf{Cross-lingual RE.}
Cross-lingual RE has been approached through multilingual encoders such as mBERT and XLM-RoBERTa~\cite{conneau2020unsupervised} trained on source-language labels, through annotation projection over aligned parallel corpora, and through machine translation of the training set~\cite{faruqui2015multilingual}, which is the approach used in this work.\\

\noindent \textbf{LLMs for Information Extraction.}
Wei et al.~\cite{wei2023zeroshot} examine zero-shot prompting of LLMs on NER and RE and report that prompt-only models trail fine-tuned baselines on standard benchmarks.
Parameter-efficient methods reduce the gap.
LoRA~\cite{hu2021lora} learns low-rank updates over frozen weights, and QLoRA~\cite{dettmers2023qlora} pairs low-rank updates with 4-bit quantization, allowing fine-tuning of 30B-scale models on a single A100.
Our QLoRA configuration follows the original recipe with the standard target modules and rank 32.\\

\noindent \textbf{Romanian NLP Resources.}
Two monolingual BERT-style models are available for Romanian: 
1) BERT-base-Romanian by Dumitrescu et al.~\cite{dumitrescu2020rombert}, and 
2) RoBERT by Masala et al.~\cite{masala2020robert}.
The RoNEC corpus~\cite{dumitrescu2020ronec} is the standard Romanian NER benchmark.
No public Romanian RE dataset of comparable scale to SemEval-2010 Task 8 has been released, and the translated dataset introduced here is intended to address this absence.

\section{Methodology}
\label{sec:method}
Our methodology maps directly onto the three research questions. 
The dataset construction and its validation address $\bm{Q_1}$, since the translation step is where performance can be lost.
The zero-shot and few-shot inference configurations address $\bm{Q_2}$.
The QLoRA fine-tuning, together with the encoder baselines that put its results in context, addresses $\bm{Q_3}$.

\subsection{Dataset Construction}

The English source data is SemEval-2010 Task 8~\cite{hendrickx2010semeval}, with 8\,000 training and 2\,717 test sentences.
Each sentence contains two entities tagged with \texttt{<e1>} and \texttt{<e2>}; the gold label is one of nine directional relations or \texttt{Other}, with direction encoded as in \texttt{Cause-Effect(e1,e2)}.

The Romanian version is produced by translating each sentence with Claude Haiku through the Anthropic API.
The translation prompt requires the model to preserve the four entity tags, keep them in the original order, and write idiomatic Romanian.
A validation pass discards translations that miss a tag, contain unbalanced markers, or produce empty entity spans.
After validation, 7\,871 training and 2\,664 test examples remain, corresponding to a retention rate of 98.4\% and 98.0\%.

To assess translation quality beyond the automatic marker check, one author manually inspected a random sample of 100 translated training sentences.
Sentence-level fluency is high: 96 of 100 translations are grammatical and read naturally in Romanian, and the original relation label remains valid in 98 of 100 cases.
Entity fidelity is lower.
In 74 of 100 examples, both marked entities are translated correctly and aligned to the right span. The remaining 26 fall into three groups: 14 where the surrounding sentence is translated but the entity inside the markers is left in English (e.g.\ \texttt{<e1>doll</e1>} in an otherwise Romanian sentence), 9 where a marker is placed on the wrong token, and 3 where the entity is mistranslated in a way that breaks the relation (e.g., \emph{grenade} rendered as the place name \emph{Granada}).
We label 12 of the 26 as severe, meaning the relation is hard or impossible to recover from, when the Romanian entity spans alone.

This pattern has different consequences for the two task formulations.
Relation classification is robust to it: the marked tokens stay in the sentence, so the model still sees them and assigns the relation, which is why Romanian classification F1-Score stays close to English.
End-to-end extraction is penalized, since the gold entity is taken from the marker, and a correct Romanian prediction is scored as wrong when the gold span is left untranslated or misplaced.
The Romanian end-to-end numbers in Section~\ref{sec:results} should therefore be read as a lower bound.
We report the dataset as machine-translated with automatic post-validation rather than as a human-quality resource, and we leave a cleaning pass over the flagged entity errors to future work.

Table~\ref{tab:dataset} lists basic statistics.
The label distribution is biased, and \texttt{Other} is the majority class.
Since translation is applied uniformly across labels, the imbalance carries over to Romanian.

\begin{table}[!htbp]
\centering
\caption{Dataset statistics after translation and validation.}
\label{tab:dataset}
\begin{tabular}{lcc}
\toprule
 & \textbf{English} & \textbf{Romanian} \\
\midrule
Train examples & 8,000 & 7,871 \\
Test examples  & 2,717 & 2,664 \\
Retention rate & 98.4\% & 98.0\% \\
Relation types & 10 & 10 \\
\bottomrule
\end{tabular}
\end{table}

\subsection{Task Formulations}

We evaluate two task formulations. In \textbf{Relation Classification}, the entity tags \texttt{<e1>} and \texttt{<e2>} are kept in the input, and the model selects one of the ten relations with its direction. In \textbf{End-to-End RE}, the entity tags are stripped, and the model has to recover both entities and the relation between them in a single generation.

\subsection{Inference Configurations}

\noindent \textbf{Zero-shot.} 
The prompt enumerates the ten relations with one-line descriptions and asks for the label and direction. \\

\noindent \textbf{Few-shot.} 
We add $k$ labeled examples to the start of the prompt, with $k \in \{1, 3, 5\}$, sampled at random from the training set in the same language as the test sentence. \\

\noindent \textbf{QLoRA fine-tuning.}
Gemma 4 31B-it is fine-tuned with QLoRA~\cite{dettmers2023qlora} in 4-bit on the combined English and Romanian training data (15\,871 examples).
The LoRA configuration uses rank 32, $\alpha = 64$, dropout 0.05, and is applied to all attention and MLP projections.
Training runs for three epochs with an effective batch size of 16, peak learning rate $2 \times 10^{-4}$ under cosine decay, and 5\% warmup.\\

\noindent \textbf{Encoder baselines.}
Four encoder models are fine-tuned on the same data.
XLM-RoBERTa~\cite{conneau2020unsupervised} base (278M) and large (560M) are trained jointly on English and Romanian and evaluated on both test sets.
BERT-base-Romanian-cased~\cite{dumitrescu2020rombert} (125M) and RoBERT-large~\cite{masala2020robert} (340M) are monolingual and are trained on the Romanian split and evaluated only on the Romanian test set.
The four markers \texttt{<e1>}, \texttt{</e1>}, \texttt{<e2>}, \texttt{</e2>} are replaced with four special tokens added to the vocabulary, and the \texttt{[CLS]} representation feeds a linear classifier over 19 directional labels, which are collapsed to 10 coarse labels at evaluation time.
Training uses a batch size of 16 or 32, depending on model size, a learning rate of $2 \times 10^{-5}$, 5 epochs, 10\% warmup, and a weight decay of 0.01.
The best checkpoint is selected by macro F1-Score on a held-out validation split (10\% of the training data), and the test set is used only for the final evaluation reported below.

\subsection{Model and Infrastructure}

Gemma 4 31B-it~\cite{gemma2026} is loaded in 4-bit through \texttt{bitsandbytes}.
The encoders are loaded in bfloat16 (BF16), a 16-bit floating-point format that keeps the exponent range of 32-bit floats while reducing memory use.
All experiments are run on a single NVIDIA A100 40GB GPU.
Differences in macro F1-Score between models are assessed with a paired bootstrap test over the test instances (10\,000 resamples).

\section{Experimental Results}
\label{sec:results}

We organize the results around the three research questions.
The translation quality assessment in Section~\ref{sec:method} together with the cross-lingual gap reported below answers $\bm{Q_1}$.
The zero-shot and few-shot results in the next two subsections answer $\bm{Q_2}$.
The QLoRA results, read against the encoder baselines and the compute cost, answer $\bm{Q_3}$.

\subsection{Relation Classification}

Table~\ref{tab:classification} reports macro F1-Score and accuracy for Relation Classification.
Gemma 4 zero-shot reaches 0.655 on English and 0.622 on Romanian.
Few-shot prompting moves the score by less than 1pp in either direction: 1-shot and 3-shot are slightly below zero-shot on English, 5-shot is slightly above, and the same holds on Romanian, where 3-shot peaks at 0.631.

\begin{table}[!htbp]
\centering
\caption{Relation Classification. Macro F1-Score and accuracy on the SemEval-2010 Task~8 test set. ``--'' indicates a monolingual Romanian model not evaluated on English. Best results in bold.}
\label{tab:classification}
\begin{tabular}{llcccc}
\toprule
 & & \multicolumn{2}{c}{\textbf{English}} & \multicolumn{2}{c}{\textbf{Romanian}} \\
\cmidrule(lr){3-4} \cmidrule(lr){5-6}
\textbf{Model} & \textbf{Params} & F1-Score & Acc & F1-Score & Acc \\
\midrule
Gemma zero-shot    & 31B  & .655 & .687 & .622 & .661 \\
Gemma few-shot-1   & 31B  & .642 & .683 & .616 & .661 \\
Gemma few-shot-3   & 31B  & .637 & .679 & .631 & .666 \\
Gemma few-shot-5   & 31B  & .660 & .686 & .617 & .663 \\
\midrule
BERT-ro-base       & 125M & --    & --    & .824 & .812 \\
XLM-R-base         & 278M & .853 & .845 & .822 & .814 \\
RoBERT-large       & 340M & --    & --    & .844 & .834 \\
XLM-R-large        & 560M & .875 & .865 & .857 & .848 \\
Gemma + QLoRA      & 31B  & \textbf{.880} & \textbf{.868} & \textbf{.865} & \textbf{.850} \\
\bottomrule
\end{tabular}
\end{table}

QLoRA fine-tuning raises macro F1-Score to 0.880 on English and 0.865 on Romanian, gains of 22.5pp and 24.3pp over zero-shot.
The cross-lingual gap narrows from 3.3pp to 1.4pp.
Per-class F1-Score is above 0.85 for most relations, with \texttt{Other} the exception at 0.71 on English and 0.67 on Romanian, which is expected given its catch-all definition.

The encoder baselines are close behind.
XLM-R-large reaches 0.875 English and 0.857 Romanian, XLM-R-base 0.853 English and 0.822 Romanian, RoBERT-large 0.844 Romanian, and BERT-base-Romanian 0.824 Romanian.
The four encoders span 3.5pp on Romanian, and the gap from the smallest encoder to QLoRA Gemma is 4.1pp on Romanian.
BERT-ro-base (125M) and XLM-R-base (278M) reach 0.824 and 0.822 on Romanian, so a smaller monolingual model matches a larger multilingual one when only the target language is in use.

\subsection{End-to-End Extraction}

End-to-End results are reported in Table~\ref{tab:e2e} under three metrics: exact match (both entities and the relation are correct), relation match (the relation type is correct without checking entity spans), and entity match (at least one of the two entity spans is correctly identified).
Absolute numbers are lower than for classification, as expected from the harder setting.

\begin{table}[!htbp]
\centering
\caption{End-to-End Extraction. Exact match, relation match, and entity match on English and Romanian.}
\label{tab:e2e}
\begin{tabular}{lccc}
\toprule
\textbf{Setting} & \textbf{Exact} & \textbf{Relation} & \textbf{Entity} \\
\midrule
\multicolumn{4}{l}{\textit{English}} \\
\midrule
Zero-shot    & .324 & .563 & .457 \\
Few-shot-1   & .371 & .580 & .503 \\
Few-shot-3   & .356 & .591 & .491 \\
Few-shot-5   & .354 & .595 & .492 \\
QLoRA        & \textbf{.719} & \textbf{.816} & \textbf{.796} \\
\midrule
\multicolumn{4}{l}{\textit{Romanian}} \\
\midrule
Zero-shot    & .285 & .537 & .412 \\
QLoRA        & \textbf{.674} & \textbf{.809} & \textbf{.751} \\
\bottomrule
\end{tabular}
\end{table}

For the English dataset, exact match goes from 0.324 in zero-shot to 0.371 with one-shot and settles around 0.355 for three- and five-shot.
Relation match stays in the 0.56-0.60 range across configurations.
The 0.20-0.25 absolute gap between relation match and exact match indicates that the errors come from entity span identification rather than from relation classification.

On the Romanian dataset, zero-shot exact match is 0.285 and relation match is 0.537, gaps of 3.9pp and 2.6pp from English.
The drop is roughly proportional across metrics, with no single metric collapsing.

Few-shot examples have a larger effect on End-to-End Extraction than on Relation Classification.
The English exact match improves by 4.7pp from zero-shot to one-shot, while the relation match moves only 1.7pp in the same direction.
The examples act mostly as output format anchors.

QLoRA fine-tuning produces a large jump on this task.
Exact match rises from 0.324 to 0.719 on English and from 0.285 to 0.674 on Romanian, gains of 39.5pp and 39.0pp over zero-shot.
Relation match reaches 0.816 English and 0.809 Romanian, and entity match 0.796 English and 0.751 Romanian.
The improvement is much larger than the 22 to 24pp seen on classification, which is consistent with the model learning both the JSON output format and the entity span conventions that it cannot infer from the prompt alone.
The cross-lingual gap on exact match is 4.5pp, wider than the 1.4pp on classification, since end-to-end extraction depends on entity boundaries that are more sensitive to translation than relation labels.

\subsection{Per-Relation Cross-lingual Behavior}

The aggregate cross-lingual gap hides uneven behavior across relation types.
Table~\ref{tab:perrel} reports per-relation F1-Score for the two strongest models, QLoRA Gemma and XLM-R-large, on both languages.
The drop from English to Romanian is not uniform.
The \texttt{Instrument-Agency} relation decreases the most in F1-Score when using XLM-R-large, from 0.86 to 0.80, and is also one of the weakest classes in absolute terms.
It has the smallest support of the directional relations (150 test instances), so the translation of a handful of examples affects its score more than for frequent classes.
The \texttt{Product-Producer} relation also decreases on Romanian under both models, and \texttt{Other} stays the weakest class in absolute terms.

\begin{table}[!htbp]
\centering
\caption{Per-relation F1-Score for the two strongest models. The cross-lingual drop concentrates in \texttt{Instrument-Agency}, \texttt{Product-Producer}, and \texttt{Other}.}
\label{tab:perrel}
\setlength{\tabcolsep}{4pt}
\begin{tabular}{lcccc}
\toprule
 & \multicolumn{2}{c}{\textbf{Gemma+QLoRA}} & \multicolumn{2}{c}{\textbf{XLM-R-large}} \\
\cmidrule(lr){2-3} \cmidrule(lr){4-5}
\textbf{Relation} & English & Romanian & English & Romanian \\
\midrule
Cause-Effect        & .92 & .93 & .95 & .94 \\
Instrument-Agency   & .86 & .85 & .86 & .80 \\
Product-Producer    & .91 & .87 & .89 & .87 \\
Content-Container   & .92 & .92 & .91 & .91 \\
Entity-Origin       & .88 & .87 & .88 & .86 \\
Entity-Destination  & .93 & .93 & .94 & .93 \\
Component-Whole     & .90 & .89 & .87 & .85 \\
Member-Collection   & .89 & .86 & .88 & .87 \\
Message-Topic       & .86 & .86 & .91 & .90 \\
Other               & .71 & .67 & .66 & .65 \\
\bottomrule
\end{tabular}
\end{table}

Several relations are stable across languages. \texttt{Cause-Effect}, \texttt{Entity-Destination}, and \texttt{Content-Container} stay within 1pp of their English score on both models, and \texttt{Cause-Effect} is slightly higher on Romanian under QLoRA Gemma. These are relations expressed through explicit lexical cues (verbs of causation, motion, or containment) that survive translation, whereas \texttt{Instrument-Agency} and \texttt{Product-Producer} rely on more implicit argument roles that the translation step can blur. The \texttt{Other} class is the weakest in both languages regardless of model, which follows from its role as a catch-all for unrelated entity pairs.

\subsection{Discussion}

Table~\ref{tab:gap} summarizes the cross-lingual gap on classification across all settings with both languages. Prompt-only configurations vary between 0.6pp and 4.3pp without a clear pattern in the number of shots $k$. Fine-tuning produces a smaller and more stable gap: 3.1pp for XLM-R-base, 1.8pp for XLM-R-large, and 1.4pp for QLoRA Gemma. Joint English+Romanian training narrows the asymmetry across both encoder and decoder architectures.

\begin{table}[!htbp]
\centering
\caption{Cross-lingual gap (English $-$ Romanian) in macro F1-Score for relation classification across all configurations. Monolingual models are excluded since they have no English evaluation.}
\label{tab:gap}
\begin{tabular}{lccr}
\toprule
\textbf{Setting} & \textbf{English} & \textbf{Romanian} & \textbf{Gap (pp)} \\
\midrule
Gemma zero-shot    & .655 & .622 & 3.3 \\
Gemma few-shot-1   & .642 & .616 & 2.6 \\
Gemma few-shot-3   & .637 & .631 & 0.6 \\
Gemma few-shot-5   & .660 & .617 & 4.3 \\
XLM-R-base         & .853 & .822 & 3.1 \\
XLM-R-large        & .875 & .857 & 1.8 \\
Gemma + QLoRA      & .880 & .865 & 1.4 \\
\midrule
Prompt-only avg    & .649 & .622 & 2.7 \\
\bottomrule
\end{tabular}
\end{table}

The few-shot results are flat.
Prepending one, three, or five examples changes macro F1-Score by less than 1pp relative to zero-shot, in either direction. 
Similar behavior has been reported in earlier work on instruction-tuned LLMs: the model already encodes the task through its instruction tuning, and additional examples mostly add context length and randomness.

QLoRA changes this for the LLM.
The 22.5pp gain on English and 24.3pp on Romanian over zero-shot indicates that the relation extraction capability of Gemma 4 is not the limiting factor in the prompt-only setting; the limiting factor is output format and calibration to the SemEval-2010 label set.
The larger gain on Romanian is consistent with the model having seen less RE-relevant Romanian supervision during pretraining, leaving more room for adaptation.

The encoder baselines stay close to the LLM across the size range.
XLM-RoBERTa-large reaches an F1-Score of 0.875 English and 0.857 Romanian, within 0.5-0.9pp of QLoRA Gemma.
RoBERT-large at 340M reaches 0.844 Romanian, 2.1pp below the LLM.
BERT-base-Romanian at 125M reaches 0.824 Romanian, 4.1pp below the LLM.
The F1-Score spread on Romanian across model sizes that differ by more than two orders of magnitude is under 4.5pp, which indicates that pretraining on the target language compensates for less parameter capacity when only that language is in use.

A paired bootstrap test (10\,000 resamples) puts these differences in perspective.
On the English dataset, the 0.5pp gap between QLoRA Gemma and XLM-R-large is not significant ($p = 0.23$, 95\% confidence interval (CI) $[-0.8, 1.6]$pp): the 31B model and the 560M encoder are statistically indistinguishable.
For the Romanian dataset, the same comparison is also not significant ($0.9$pp, $p = 0.09$, CI $[-0.4, 2.2]$pp), so on both languages the strongest encoder matches the LLM within noise. Only the smaller RoBERT-large falls significantly below QLoRA Gemma on Romanian ($2.1$pp, $p = 0.001$, CI $[0.8, 3.5]$pp). The advantage of the LLM, where it exists at all, is at most a couple of points.

End-to-end extraction behaves differently from classification.
Prompt-only Gemma 4 reaches only 0.324 exact match on English under zero-shot, against 0.655 F1-Score on classification.
Roughly half of the entities are recovered (entity match 0.45 on English, 0.41 on Romanian), but combining correct entities with the correct relation in a single generation is harder than either step in isolation.
QLoRA changes this sharply: exact match rises to 0.719 English and 0.674 Romanian, a gain of around 39pp.
This is where the LLM earns its place.
Encoder classifiers do not apply directly to end-to-end extraction, since they require a separate entity recognition stage, whereas the fine-tuned LLM handles entity identification and relation typing in one pass.
The compute argument that favors encoders for classification does not carry over to this setting.

\subsection{Compute Cost}

Table~\ref{tab:compute} lists training and inference costs for the fine-tuned models on the same A100 40GB.
Encoder training runs in 15 to 30 minutes; QLoRA on Gemma 4 31B takes 5h13min.
Inference over both test sets (5,381 examples) takes 1-2 seconds for the encoders and around 3.5 hours for QLoRA Gemma.
Inference memory is 1-3 GB for the encoders and 25 GB for Gemma in 4-bit.
The accuracy advantage of QLoRA Gemma over the strongest encoder is 0.5pp on English and 0.8pp on Romanian; against RoBERT-large on Romanian, it is 2.1pp.
For deployments serving Romanian RE in isolation, fine-tuned RoBERT-large is the more practical option.

\begin{table}[!htbp]
\centering
\caption{Approximate compute cost across approaches on a single NVIDIA A100 40GB. Inference time is for both test sets ($\sim$5.4K examples).}
\label{tab:compute}
\begin{tabular}{lcccc}
\toprule
 & \textbf{Params} & \textbf{Train} & \textbf{Infer} & \textbf{Mem} \\
\midrule
BERT-ro-base    & 125M & 15 min   & 1 s    & 1 GB \\
XLM-R-base      & 278M & 20 min   & 1 s    & 2 GB \\
RoBERT-large    & 340M & 30 min   & 2 s    & 2 GB \\
XLM-R-large     & 560M & 30 min   & 2 s    & 3 GB \\
Gemma + QLoRA   & 31B  & 5h 13min & $\sim$3.5 h & 25 GB \\
\bottomrule
\end{tabular}
\end{table}

The LLM remains a reasonable choice when the same model is also used for other tasks or for open-ended generation, including the End-to-End RE formulation, where QLoRA fine-tuning more than doubles exact match over zero-shot.

\section{Limitations and Future Work}
\label{sec:future}

Our study has several limitations.
The Romanian dataset is produced by automatic translation with automatic post-validation, and the manual inspection in Section~\ref{sec:method} shows that 26\% of a 100-sentence sample has entity-level translation errors, 12\% of them severe.
This affects End-to-End RE in particular, where the reported Romanian exact and entity match scores are a lower bound, and a cleaning pass over the flagged entity errors remains future work.
We evaluate a single LLM (Gemma 4 31B) and four encoder baselines; other open-weight models such as Llama 3 or Qwen 2.5 may behave differently.
Each configuration is run with a single seed, and per-seed variance for the few-shot prompts has not been measured, which may explain part of the inconsistency across the number of shots.
Finally, the experiments cover one benchmark (SemEval-2010 Task 8) with ten relation types, so the conclusions may not transfer directly to datasets with a larger or domain-specific relation inventory.

Beyond addressing these limitations, two directions extend this work.
Symbolic post-processing of the generated tuples through type constraints derived from the relation schema can filter outputs that violate the expected argument types, which connects to our broader work on hybrid neuro-symbolic methods for cross-lingual information extraction.
Evaluating additional open-weight models would also test whether the encoder-versus-LLM trade-off observed here generalizes beyond Gemma 4.

\section{Conclusions}

We studied cross-lingual relation extraction for Romanian by translating SemEval-2010 Task 8 from English and evaluating Gemma 4 31B against four encoder baselines under Relation Classification and End-to-End RE.

Automatic translation preserves most of the relational signal.
The cross-lingual gap in Relation Classification is 3pp to 5pp in prompt-only settings and narrows to 1.4pp after fine-tuning, although manual inspection shows that entity-level translation errors set a lower bound on the End-to-End RE results --- answering $\bm{Q_1}$.
Zero-shot and few-shot prompting are not enough on their own.
Few-shot examples change macro F1-Score by less than 1pp over zero-shot, and prompt-only Gemma stays far below the fine-tuned models on both tasks --- answering $\bm{Q_2}$.
QLoRA fine-tuning on the translated data is effective, adding more than 22pp of macro F1-Score on Relation Classification and around 39pp of exact match on End-to-End RE in both languages --- answering $\bm{Q_3}$.

On Relation Classification, the strongest encoder reaches within 0.5 to 0.9pp of the 31B model and the difference is not significant on English, so a model 50 to 250 times smaller is the more economical choice.
On End-to-End RE, where entities are not given and encoder classifiers do not apply directly, the fine-tuned LLM has a clear advantage and justifies its cost.
For Romanian relation extraction, task-specific encoder models remain highly competitive despite being far smaller than instruction-tuned LLMs, and the case for a large model is strongest when the task requires open-ended generation.

\section*{Use of Generative AI}

The authors used Claude (Anthropic) to assist with grammar and spelling checks on the manuscript text. All technical content, experimental design, analysis, and conclusions are the authors' own. The Claude Haiku model is also a methodological component of this work, used for translating the SemEval-2010 Task~8 dataset from English to Romanian as described in Section~\ref{sec:method}.

\bibliographystyle{IEEEtran}
\bibliography{references}

\end{document}